# Towards Reliable Sequential Object Picking in Clutter: The Runner-up Solution to RGMC 2025

Wei Yu[1,*], Xidan Zhang[1,*], Ziyi Zheng[1,*], Weijie Kong[1] and Huixu Dong[1,†]

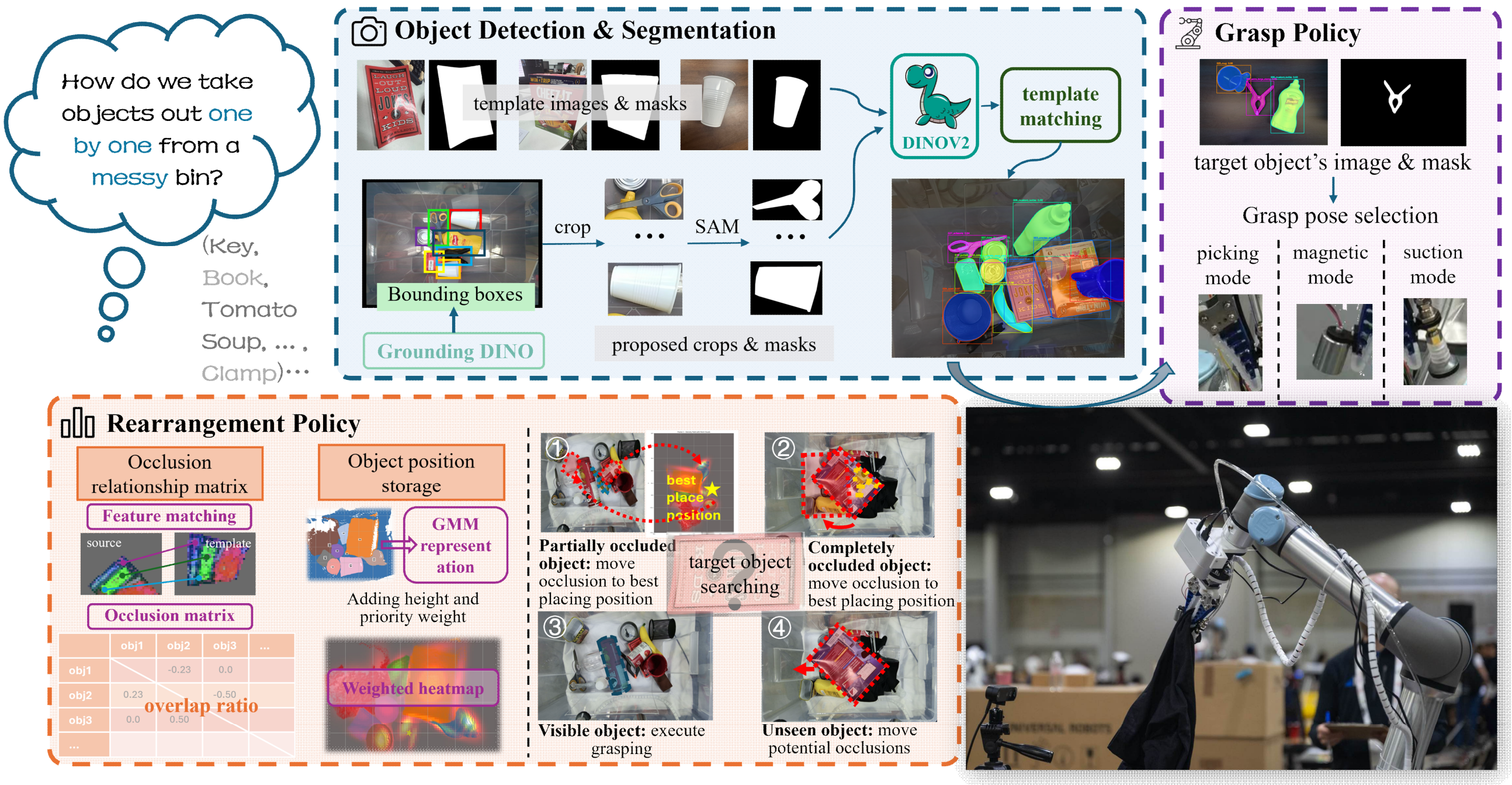

Fig. 1: **Overview of our pipeline for sequential object picking from cluttered bins,** which encompasses the object detection and segmentation process **(top-left)**, rearrangement policy **(bottom)** and grasp policy **(top-right)**.

## I. Introduction

As a long-standing challenge in robotic manipulation, stable and efficient grasping in cluttered environments is of great importance in industrial settings. While recent studies[1-5] have achieved relatively high success rates in grasping from clutter, there remain few mature solutions for more demanding tasks such as sequential object search and sorting.

This work addresses sequential object picking in cluttered environments based on the Cluttered Environment Picking Benchmark (CEPB)[6] and presents our solution to the Pick-in-Clutter track of the 10th *Robotic Grasping and Manipulation Competition (RGMC)* at ICRA 2025. The task poses several key challenges. First, it requires robust and collision-aware grasping with high success rates across a diverse set of objects, including both rigid and deformable ones. Second, it demands efficient search for target objects, which places stringent requirements on the decluttering and searching strategies of the solution.

[1] School of Mechanical Engineering, Zhejiang University, Hangzhou, China. Emails: {wei_yu, xidanzhang, 12325091, wjkong}@zju.edu.cn
* Equal contribution
† Corresponding author

To address the above challenges, we design an integrated hardware-software pipeline that combines object recognition, decluttering, and multi-modal grasping. The main contributions include the hardware design of a multifunctional gripper and novel representations for object distribution and occlusion relationships in cluttered space. This pipeline enables efficient recognition, search, and sequential grasping of objects in clutter, demonstrating strong performance in both laboratory tests and competition scenarios, and ultimately achieving second place in the Pick-in-Clutter track of the RGMC 2025.

## II. Method

### A. Hardware Setup

Our hardware system consists of a UR5e robotic arm, a robotic gripper, and a wrist-mounted Orbbec Femto Bolt camera. To achieve stable grasping performance across the diverse object categories in the benchmark, we designed an adaptive gripper that integrates a Fin-ray soft gripper, an electromagnet, and a suction cup. This design enables three grasping modalities: mechanical grasping, magnetic picking, and suction-based picking. The hardware system also includes the corresponding electronic controller and an air pump. The complete hardware setup used in the competition is shown in Figure 2.

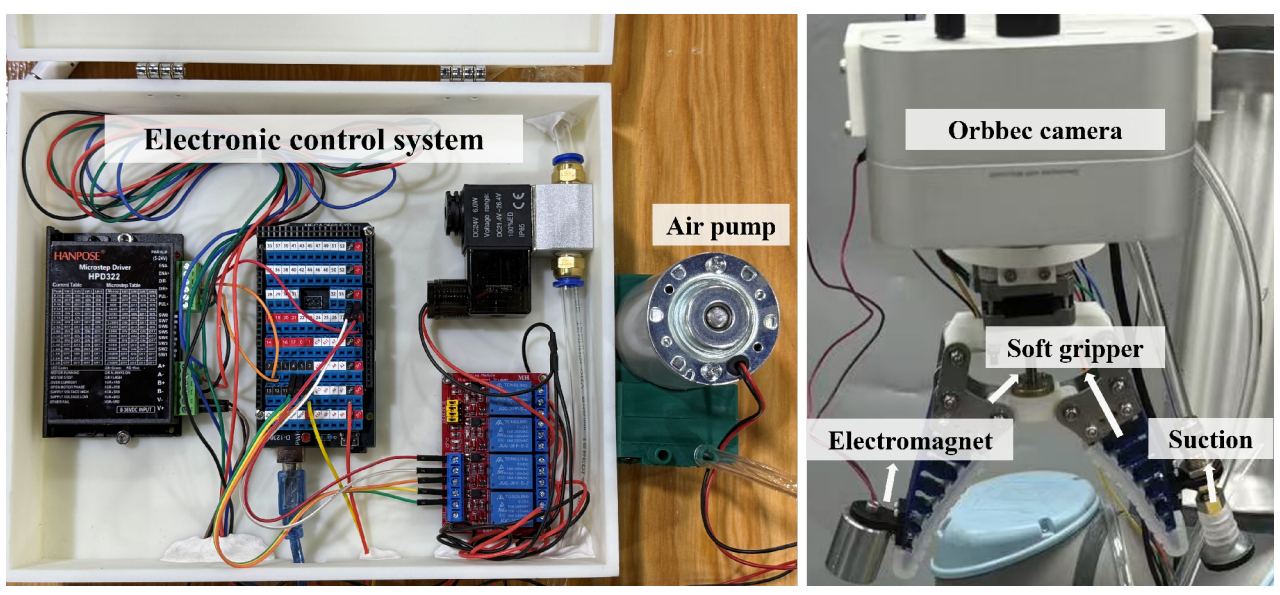

Fig. 2: Hardware setup consisting of the UR5e robotic arm, adaptive gripper, and wrist-mounted camera.

## B. Object Detection and Segmentation

To achieve real-time detection and segmentation of individual objects in clutter, we adopt NIDS-Net[7], a recently proposed framework for novel instance detection and segmentation. NIDS-Net combines object proposal generation, embedding construction for both templates and candidate regions, and embedding matching for instance assignment. Leveraging large vision models such as Grounding DINO[8] and the Segment Anything Model (SAM) [9], it produces accurate proposals with bounding boxes and masks. The resulting object masks are then used in our decision-making and grasping pipeline.

## C. Rearrangement Policy

The main challenge in sequentially extracting target objects from clutter lies in removing occluding objects and locating objects that are not immediately visible. To tackle this, we model inter-object occlusions and spatial distributions. For occlusion, we build a matrix recording pairwise occlusion ratios, where exposed regions and complete templates are compared via DINOv2[10] feature matching. In parallel, spatial memory encodes potential object distributions, with updated point clouds converted into a Gaussian Mixture Model (GMM) representation. Considering object priority from height and sequence order, we assign dynamic weights, and project the weighted GMM along the z-axis to produce a heatmap, guiding object rearrangement and search strategies.

For each target object, our strategy considers three cases. If the object is currently visible, it is directly grasped. If it is partially or fully occluded, the occluding objects are relocated to suitable positions selected from the current heatmap, with the goal of minimizing additional occlusion in the clutter. If the object has never been observed, random object relocation and search are first performed in the most probable regions.

## D. Grasp Policy

For each object to be manipulated, whether a target or an occluding object, we select an appropriate grasping modality from three options—gripper grasping, suction, or magnetic pickup—based on its category. For rigid-surfaced objects, suction is most commonly applied; for metallic items, magnetic pickup is generally preferred; and for flexible or irregularly shaped objects, we primarily rely on the gripper modality. Each modality has a dedicated grasp pose generation strategy, as shown in Figure 3. For suction-based modalities, candidate points are obtained by plane fitting on the object's point cloud and scored according to margin area, distance to the center of gravity (COG) for object, and the angle between the surface normal and the z-axis, yielding optimal candidates. For gripper-based grasping, we leverage AnyGrasp[11] to generate grasp poses from the point cloud. Since these methods do not inherently account for collisions with surrounding objects or the bin, we further perform collision-avoidance filtering on the candidate poses. The final selected grasp pose is then passed to the arm trajectory planner for execution.

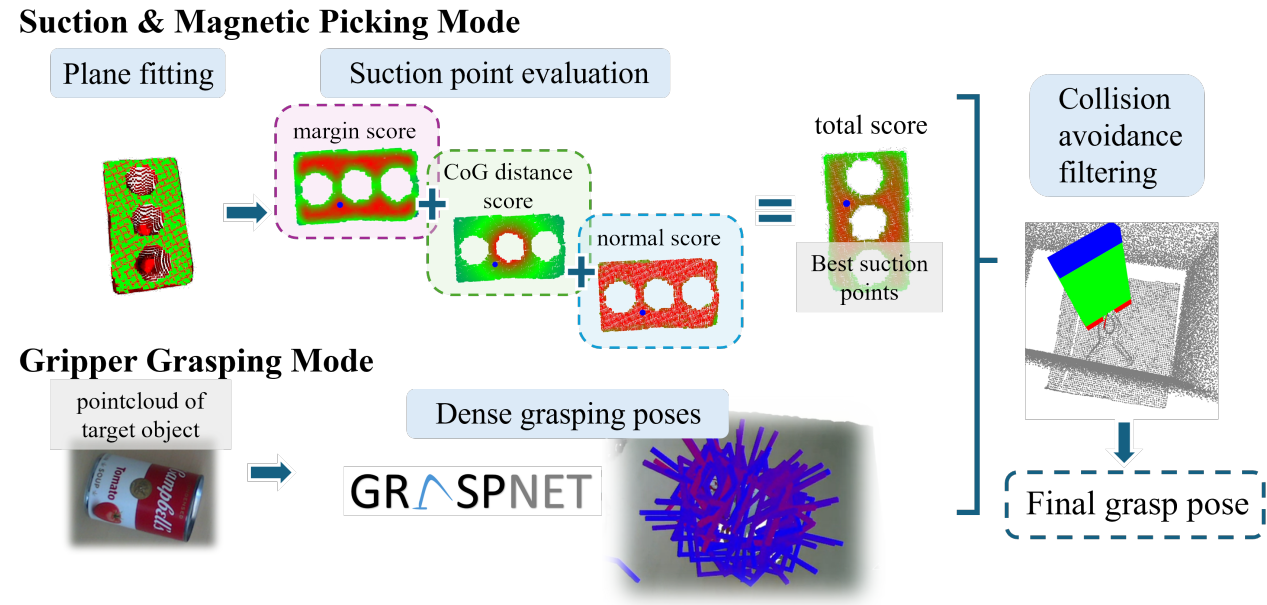

Fig. 3: Overview of grasp pose generation pipeline across modalities.

# III. Experiments

We conducted a series of repeated experiments to validate our pipeline in terms of sub-functionalities such as object recognition and grasping, as well as its overall performance in sequential object extraction from clutter. Upon receiving the grasping sequence via ROS topics, performance was measured by counting the number of objects sequentially retrieved within a fixed time window, with success rate defined as the ratio of successfully extracted objects to the total number of objects in the bin. Our evaluation included both laboratory testing and real competition scenarios. In the laboratory, we performed 10 rounds of full trials, achieving an average task completion success rate of 68%. In the competition, we conducted three trials with different target sequences, where the numbers of *Picked Objects* were 2–1–11, and the numbers of *Picked Objects in order* were 1–1–8, respectively. Apart from the program errors in the first and second rounds that prevented the robot from executing actions properly, an analysis of the third-round experiment and our routine laboratory tests reveals that the most frequent failure cases include:

- False or failed detection: confusion in recognizing visually similar objects (e.g., black gloves versus black T-shirts); poor point cloud quality for transparent objects leading to inaccurate grasp poses.
- Spatial constraints: objects located in confined spaces (e.g., at the corner of the bin); especially for small-sized objects, the likelihood of collisions between the gripper and the bin increases, resulting in grasp failures.

These issues can be mitigated by refining the gripper design to make it more dexterous and better adapted to cluttered environments, improving the robustness of object recognition, and enhancing the handling of transparent objects. Moreover, to improve the efficiency of sequential grasping, strategies such as introducing in-bin stirring or flipping operations can be integrated. These constitute the main directions of our future work.

## REFERENCES

[1] H. Dong, E. Asadi, G. Sun, D. K. Prasad and I.-M. Chen, "Real-time robotic manipulation of cylindrical objects in dynamic scenarios through elliptic shape primitives," *IEEE Transactions on Robotics*, vol. 35, no. 1, pp. 95–113, Feb. 2019.

[2] J. Li, K. Zhu, G. Lu, I.-M. Chen and H. Dong, "Construction of a multiple-DOF underactuated gripper with force-sensing via deep learning," in *Proc. Robotics: Science and Systems (RSS)*, Delft, Netherlands, Jul. 2024.

[3] Y. Ma, W. Yu, Z. Su, X. Zhang and H. Dong, "SID: Sliding into Distribution for Robust Few-Demonstration Manipulation," arXiv preprint arXiv:2605.13428, May 2026.

[4] B. Wang and H. Dong, "Bin packing optimization via deep reinforcement learning," *arXiv preprint arXiv:2403.12420*, 2024.

[5] S. D'Avella, A. M. Sundaram, W. Friedl, P. Tripicchio and M. A. Roa, "Multimodal grasp planner for hybrid grippers in cluttered scenes," *IEEE Robotics and Automation Letters*, vol. 8, no. 4, pp. 2030–2037, Apr. 2023.

[6] S. D'Avella, M. Bianchi, A. M. Sundaram, C. A. Avizzano, M. A. Roa, and P. Tripicchio, "The Cluttered Environment Picking Benchmark (CEPB) for Advanced Warehouse Automation: Evaluating the Perception, Planning, Control, and Grasping of Manipulation Systems," *IEEE Robotics & Automation Magazine*, vol. 31, no. 4, pp. 45–58, Dec. 2024.

[7] Y. Lu, J. J. P. Jaykumar, Y. Guo, N. Ruozzi, and Y. Xiang, "Adapting Pre-Trained Vision Models for Novel Instance Detection and Segmentation," *arXiv preprint arXiv:2405.17859*, 2024.

[8] S. Liu, Z. Zeng, T. Ren, F. Li, H. Zhang, J. Yang, C. Li, J. Yang, H. Su, J. Zhu, et al., "Grounding DINO: Marrying DINO with Grounded Pre-Training for Open-Set Object Detection," *arXiv preprint arXiv:2303.05499*, 2023.

[9] A. Kirillov, E. Mintun, N. Ravi, H. Mao, C. Rolland, L. Gustafson, T. Xiao, S. Whitehead, A. C. Berg, W.-Y. Lo, P. Dollár, and R. Girshick, "Segment Anything," *arXiv:2304.02643*, 2023.

[10] M. Oquab, T. Darcet, T. Moutakanni, H. V. Vo, M. Szafraniec, V. Khalidov, P. Fernandez, D. Haziza, F. Massa, A. El-Nouby, R. Howes, P.-Y. Huang, H. Xu, V. Sharma, S.-W. Li, W. Galuba, M. Rabbat, M. Assran, N. Ballas, G. Synnaeve, I. Misra, H. Jegou, J. Mairal, P. Labatut, A. Joulin, and P. Bojanowski, "DINOv2: Learning Robust Visual Features without Supervision," *arXiv:2304.07193*, 2023.

[11] H.-S. Fang, C. Wang, H. Fang, M. Gou, J. Liu, H. Yan, W. Liu, Y. Xie, and C. Lu, "AnyGrasp: Robust and Efficient Grasp Perception in Spatial and Temporal Domains," *IEEE Transactions on Robotics (T-RO)*, 2023.